\pdfoutput=1

\documentclass{scrartcl}
\usepackage[utf8]{inputenc}

\title{Gzip versus bag-of-words for text classification}
\subtitle{}
\author{Juri Opitz \\ \texttt{\normalsize opitz.sci@gmail.com} }
\date{}

\usepackage{comment}
\usepackage{booktabs}
\usepackage{url}
\usepackage{adjustbox}
\usepackage{xspace}
\usepackage{multicol}
\usepackage{appendix}
\usepackage{amsmath}
\usepackage{hyperref}
\usepackage{xcolor, color, colortbl}
\usepackage{listings}

\usepackage{arydshln}
\definecolor{Gray}{gray}{0.94}

\begin{document}

\maketitle

\newcommand{\gzip}{\textsc{gzip}\xspace}
\newcommand{\simple}{\textsc{BowDist}\xspace}
\newcommand{\simpletr}{\textsc{BowTrain}\xspace}

\begin{abstract}
The effectiveness of compression in text classification (`\gzip') has recently garnered lots of attention. In this note we show that `bag-of-words' approaches can achieve similar or better results, and are more efficient. 
\end{abstract}

\section{Introduction}

KNN is a simple classifier that uses distance measurements between data points: For a given testing point, we calculate its distance to every other point from some labeled training set and check the labels of the $K$ closest points (i.e., the \textit{\textbf{K}-\textbf{N}eirest} \textbf{N}eighbors), predicting the label that we observe most frequently. 

Hence, it is straightforward to build a general text classifier from KNN, if we can equip it with a sensible distance measure between documents. Interestingly, recent findings \cite{jiang-etal-2023-low} suggest that we can exploit compression to assess the distance of two documents, by comparing their individual compression lengths to the length of their compressed concatenation (we call this measurement \gzip). With this approach, \cite{jiang-etal-2023-low} show strong text classification performance across different data sets, sometimes achieving higher accuracy than trained neural classifiers such as BERT \cite{devlin-etal-2019-bert}, especially in scenarios where only few training data are available. 

Against this background, it is not surprising that \gzip has quickly attracted lots of attention.\footnote{In only a few days, the \gzip github repository (\url{https://github.com/bazingagin/npc_gzip}) attracted more than 1.5k `stars', and twitter micro blogs on the \gzip paper that appeared at ACL 2023 got several thousands of `likes'.} However, we find that there is more to the story:

\begin{enumerate}
    \item After correcting optimistic results in \cite{jiang-etal-2023-low}, we observe some generalization loss of \gzip distance in KNN, affecting mostly its competitiveness in scenarios where sufficient training data are available. 
    
    \item Simple `bag-of-words' approaches are effective. This holds true both when building a distance measure for KNN (where bag-of-word distance can outperform \gzip), and even more so we when we train a linear bag-of-words classifier.
\end{enumerate}

\section{Experiments}

\begin{figure}
\centering
 \begin{minipage}{0.50\textwidth}
 \footnotesize
\begin{lstlisting}
def string2set(string):
    string = string.replace(".", " ")
    string = string.replace(",", " ")
    string = string.replace("?", " ")
    string = string.replace("!", " ")
    string = string.lower()
    ls = string.split()
    ls = [t for t in ls if len(t) > 3]
    return set(ls)

def distance(string1, string2):
    set1 = string2set(string1)
    set2 = string2set(string2)
    x = len(set1.intersection(set2))
    y = len(set1.union(set2))
    return 1 - x / y
\end{lstlisting}
\end{minipage}
\caption{Code for \simple. \texttt{string2set} can also be viewed as a (lossy) compression and the distance of two strings \texttt{a} and \texttt{b} can be re-written in a `compressor'-style to~ \texttt{2 - (|compress(a)| + |compress(b)|)/|compress(concat(a, b))|}, where \texttt{|x| = len(x)}.}
\label{fig:simple}
\end{figure}

\paragraph{\textbf{\textsc{\textmd{\simple}}} for a simple KNN approach} is shown in Figure \ref{fig:simple}.\footnote{Full code for experiments can be found here: \url{https://github.com/flipz357/npc_gzip_exp}.} We apply string modifications such as removing dots and commas as well as lower-casing and token splitting at white space. Then we filter out very short tokens, since we assume that they tend to not contribute (so much) to the core of a text, an assumption that has also been previously exploited \cite{2007:phd-Ana-Cardoso-Cachopo}. Finally, to compare two strings, we apply a simple bag-of-tokens set-distance based on the Jaccard index that dates back to at least 1884 \cite{gilbert1884finley}. 

\paragraph{\textbf{\textsc{\textmd{\simpletr}}} is a trained bag-of-words approach} that is created by i) using a vectorized bag-of-words representation, where dimensions indicate counts of specific words, and ii) learning class-associated weight vectors from some training data. We use popular tfidf-normalized word count vectors (discounting words that occur in many documents). For learning the weight vector we use a default linear classifier with regularization.\footnote{LinearSVC classifier and vectorizers are taken from the \texttt{scikit-learn} package.}

\subsection{Data setup, baselines, and evaluation}

\paragraph{Data sets and baselines} are mainly based on \cite{jiang-etal-2023-low}'s study. We use 20News, Ohsumed (Omed), Reuters-8 (R8), Reuters-52 (R52). KinyarwandaNews (KinNws), KirundiNews (KirNews), DengueFilipino (DFilipino), SwahiliNews (SwNews). For few shot experiments, we also report results on large data sets such as DBpedia, YahooAnswers and AGNews, and on an additional Twitter sentiment classification task with balanced class distribution. Since in DFilipino, it appears that testing and training data are identical\footnote{Presumably due to a data loading issue, c.f.: \url{https://github.com/bazingagin/npc_gzip/issues/13}} we report results on DFilipino in brackets. Our main baseline is \gzip, which we re-calculate for most experiments (see next paragraph). Same as \cite{jiang-etal-2023-low}, we mostly use K=2 in the KNN. Where possible, we also show results from BERT fine-tuning, extracting any scores from \cite{jiang-etal-2023-low}. Experiments are run on a laptop with i5-8350 CPU and 16GB memory.

\paragraph{KNN evaluation} reported in \cite{jiang-etal-2023-low} is \textit{optimistic}. This is because of the applied tie-breaking strategy needed for handling cases where we end up with a set of neighbors among which we cannot determine exactly one most frequent label. So in case we have multiple options, including the right one, \cite{jiang-etal-2023-low} select the latter. This has generated some discussion\footnote{E.g., see  \url{https://kenschutte.com/gzip-knn-paper/}.}, since at inference we usually do not know the correct label. Therefore, we report \textit{realistic} evaluation scores that are achieved by two options: \begin{enumerate}
    \item To best compare with \cite{jiang-etal-2023-low}, who mainly use optimistic K=2, we (re-)calculate realistic scores by taking the mean of the optimistic score (if there is a tie involving the correct label, pick correct) and the pessimistic score (if there is a tie involving the correct label, pick incorrect). The realistic accuracy can be viewed as the expected accuracy when using a random guesser to break ties in KNN (see Appendix \ref{app:accuracy}).
    \item Another realistic score is calculated by using K=1, which is equivalent to K=2 when applying a tie breaking strategy that is informed by distances. We denote these models \gzip-1 and \simple-1. 
\end{enumerate} 

\subsection{Results with full-training data}

\newcolumntype{g}{>{\columncolor{Gray}}r}
\begin{table}
    \centering
    \scalebox{0.835}{
    \begin{tabular}{lrrrrrrrrg}
    \toprule
   & 20News & KinNews	& KirNews	& (DFilipino) & SwNews & Omed & R8 & R52 & \textbf{mean} \\
   \cmidrule{2-10}
        \simpletr & 0.853 & 0.889 &  0.899 & (0.991)& 0.899& 0.681 & 0.974 & 0.949 & 0.878 \\
        train BERT & 0.868 & 0.838 & 0.879 & (0.979) & 0.897 & 0.741 & 0.982 & 0.960 & 0.881 \\
        \midrule
      \gzip & 0.685 & 0.891	& 0.905	 & (0.998)	& 0.927 & 0.484 & 0.954 & 0.892 & 0.820 \\ 
     [-1.6ex] \hdashline\noalign{\vspace{\dimexpr 2.1ex-\doublerulesep}}
        \simple  &  0.704 & 0.893 & 0.930 & (0.999) & 0.921 &0.518 & 0.961 & 0.915 & 0.835 \\ [-1.6ex] \hdashline\noalign{\vspace{\dimexpr 2.1ex-\doublerulesep}}
        \gzip & 0.563  & 0.828  &  0.807   & (0.851) &  0.842 & 0.331 & 0.905 & 0.833 & 0.730 \\
        \simple  & 0.578 &0.825 & 0.840 & (0.885) & 0.834 & 0.364 & 0.919 & 0.850 & 0.744\\ 
                \midrule
        \gzip-1 & 0.607 &  0.835 &   0.858  &  (0.999)  &  0.850 & 0.362 & 0.915 & 0.851 & 0.754 \\
        \simple-1 & 0.619 & 0.845 & 0.870 & (0.997) & 0.841 & 0.385 & 0.930 & 0.867 & 0.765\\
        \bottomrule
    \end{tabular}}
    \caption{Accuracy results using full training data. Dashed lines represent KNN evaluation scores using optimistic tie-breaking. Mean excludes DFilipino.}
    \label{tab:fulldat_results}
\end{table}

\paragraph{Optimistic KNN results} in Table \ref{tab:fulldat_results}\footnote{For Omed, R8, and R52, scores for \gzip differ slightly from the scores reported in \cite{jiang-etal-2023-low}, but they align with other reproductions (c.f., \url{https://kenschutte.com/gzip-knn-paper2/}, Section 3.).} (dashed lines) show that \simple performs better than \gzip on six of seven data sets, with an average score improvement of +1.5 accuracy points.

\paragraph{Realistic results} for KNN with fair tie-breaking (Table \ref{tab:fulldat_results} non-dashed lines) show reduced performance scores of KNN with either \gzip or \simple (up to -9 accuracy points, on average). The overall trend, however, is similar: With K=2, \simple increases accuracy over \gzip by +1.4 points on average and wins on six out of seven data sets (K=1, \simple-1 vs.\ \gzip-1: +1.1). 

Rather unsurprisingly, BERT, being able to better exploit the full training data, performs better than all other methods. It is interesting, however, that \simpletr achieves almost the same performance as BERT, while being much more efficient. Next we consider the methods when fewer training data are available.

\subsection{Few-shot experiments and efficiency assessment}

It is interesting to study \gzip or \simple in a few-shot scenario, where `non-parametric' methods like KNN perhaps seem most useful.

\begin{table}
    \centering
    \scalebox{0.9}{\begin{tabular}{lrrrrr}
    \toprule
  & 20News &   KinNews	& KirNews	& (DFilipino) & SwaNews  \\
  \midrule
  \simpletr & 0.384 & 0.461 & 0.596 & (0.619) & 0.639 \\
        train BERT & - & 0.240 & 0.386 & (0.409) &  0.396 \\
        \midrule
        \gzip &  0.143 & 0.266 & 0.277 & (0.376) & 0.416 \\
        \simple  & 0.164 & 0.305 & 0.383 & (0.545) & 0.425 \\ 
        \midrule
        \gzip-1 & 0.166 & 0.340 & 0.350 & (0.456) & 0.426\\
        \simple-1 & 0.194 & 0.345 & 0.451 & (0.673) & 0.538\\
        \bottomrule
    \end{tabular}}
    \caption{5-shot evaluation, mean over five runs.}
    \label{tab:few-shot-5}
\end{table}

\paragraph{5-shot evaluation results} are displayed in Table \ref{tab:few-shot-5}: \simple-1 outperforms all other tested KNN-based methods across all data sets, including BERT and the comparable \gzip-1. Similarly, \simple consistently outperforms \gzip. Overall, however, trained bag-of-words \simpletr seems superior.

\paragraph{\{5,100\}-shot results on larger English  data} are displayed in Table \ref{tab:few-shot-100}. Again, trained bag-of-words \simpletr overall provides the best accuracy. Furthermore, we see that \simple (\simple-1) offers better performance than \gzip (\gzip-1) across all data sets, while also performing a faster distance calculations.

\begin{table}
    \centering
   \scalebox{0.9}{ \begin{tabular}{lrrrrrrr}
    \toprule
  & & \multicolumn{2}{c}{AGNews} &  \multicolumn{2}{c}{DBPedia}	& \multicolumn{2}{c}{YahooAnswers} \\ 
  & shots & acc$\uparrow$ & time$\downarrow$ & acc$\uparrow$ & time$\downarrow$ &acc$\uparrow$ &time$\downarrow$  \\
  \midrule
      \simpletr & 100 & 0.787 & [0.8s]& 0.936 & [1.9s]& 0.533 & [3.1s] \\
    \simpletr & 5 & 0.419 & $''$ & 0.732 & $''$ & 0.232 & $''$ \\
  \midrule
        \gzip & 100 &  0.549 & 244.9s & 0.748 & 57073.1s & 0.212 & 6693.8s  \\
        \simple & 100 & 0.581  & 7.9s & 0.812 & 316.2s & 0.216 & 308.1s \\
        \gzip-1 & 100 & 0.599 & $''$ & 0.779 & $''$& 0.227 & $''$ \\
        \simple-1 & 100 & 0.611 & $''$& 0.825 & $''$& 0.234 & $''$\\
        \midrule
        \gzip & 5 &  0.313 & 9.9s & 0.316 & 383.3s & 0.141 & 277.0s \\
        \simple & 5 & 0.384 & 0.6s & 0.592 & 15.8s & 0.153 & 13.3s \\
        \gzip-1 & 5 & 0.374 & $''$& 0.502 & $''$& 0.148 & $''$ \\
        \simple-1 & 5 & 0.405 & $''$& 0.638 & $''$& 0.165 & $''$\\
        \bottomrule
    \end{tabular}}
    \caption{100- and 5-shot evaluation of large data sets, mean over five runs. time: time needed for processing a complete testing set. $''$: cf.\ above for an approximate value. Runtime of \simpletr is in brackets, since \simpletr uses optimized code and requires extra time for training that is not shown here.}
    \label{tab:few-shot-100}
\end{table}

\paragraph{Twitter sentiment few-shot classification} We retrieve an additional, refined twitter sentiment classification data set with balanced class distribution of positive and negative sentiment.\footnote{Data: \url{https://github.com/cblancac/SentimentAnalysisBert/tree/main/data}. See \url{https://huggingface.co/datasets/carblacac/twitter-sentiment-analysis} for more information.} The random baseline here is simply 0.5 accuracy. Results are displayed in Table \ref{tab:few-shot-twitter}: for very low number of shots (5), both \simple and \gzip perform randomly. When more shots are available, \simple outperforms \gzip across the board with about +2 to +4 accuracy points. However, again, we see that \simpletr provides the best results with sometimes up to +10 points improvement over the next best system.

\begin{table}
    \centering
   \scalebox{0.75}{ \begin{tabular}{lrrrrrrrrrrrrr}
    \toprule
  &  \multicolumn{2}{c}{5-shot} &  \multicolumn{2}{c}{25-shot}	& \multicolumn{2}{c}{50-shot} & \multicolumn{2}{c}{100-shot} & \multicolumn{2}{c}{250-shot} & \multicolumn{2}{c}{1000-shot}\\
  &  acc$\uparrow$ & time$\downarrow$ & acc$\uparrow$ & time$\downarrow$ &acc$\uparrow$ &time$\downarrow$&acc$\uparrow$ &time$\downarrow$ &acc$\uparrow$ &time$\downarrow$ &acc$\uparrow$ &time$\downarrow$ \\
  \midrule
  \simpletr & 0.542 & [0.0s]& 0.545& [0.0s] & 0.583& [0.0s]  & 0.581& [0.0s] & 0.645& [0.0s] & 0.692 & [0.0s]  \\
  \midrule
  \gzip & 0.502 &  30.9s & 0.516 & 164.9s & 0.523 & 313.1s & 0.521 &724.3s & 0.539 & 1914.4s & 0.557 & 6894.8s \\
  \simple & 0.503 & 2.2s & 0.535 & 5.6s & 0.546 & 10.5s  & 0.554 & 18.3s & 0.561 & 54.0s & 0.594 & 262.3s \\
  \midrule 
  \gzip-1 &0.513&29.3s&0.515&183.2s&0.530&302.1s& 0.533 & 733.0s & 0.547 & 1912.8s & 0.559 & 7208.0s  \\
  \simple-1 &0.513& 2.6s& 0.53& 6.5s & 0.547 & 9.9s & 0.558 & 19.1s & 0.576 & 52.3s & 0.597 & 271.2s\\
        \bottomrule
    \end{tabular}}
    \caption{Few-shot evaluation on Twitter sentiment, mean over five runs.}
    \label{tab:few-shot-twitter}
\end{table}

\subsection{Assessing variations of our approaches}

\paragraph{\textbf{\textsc{\textmd{\simple}}} variations} In Table \ref{tab:filter_ablation}, Appendix \ref{app:hyper-param}, we study \simple in different configurations. Regarding the removal of tokens that do not exceed a length of $n$, we find that lower values of $n$ seem to work similarly well. However, when $n$ is increased too much, say $n=10$, we filter out too many tokens and the performance degrades notably (see last row `n=10' in \ref{tab:filter_ablation}). Abstaining from the removal of dots, commas and leaving case tends to result in somewhat lower performance, particularly for 20News (-2.8 accuracy points). Worst performance is achieved when inverting the token-length threshold and keeping only very short tokens (-33 points). Overall, we can conclude that \simple seems robust to different variations, if they are not too unreasonable.

\paragraph{\textbf{\textsc{\textmd{\simpletr}}} variations} We experiment with using other vector representations: binary vectors that show whether a certain words occurs, and count vectors that show how often certain words occur. Both variations perform similarly, but lag behind tfidf type vectors (see Table \ref{tab:filter_ablation} in Appendix \ref{app:hyper-param}).

\section{Conclusion and discussion}

Our findings are:

\begin{enumerate}
    \item When using KNN for text classification, we need to parameterize it with a sensible distance measure. In that aspect, we find that a `bag-of-words' (\simple) approach seems preferable to compression (\gzip), since it achieves similar or better performance. A disadvantage may be the lossiness of its representation. However, this theoretical downside seems outweighed by \simple's  greater efficiency. 
    
   \item We also find that a trained bag-of-word approach (\simpletr) achieves strong performance in almost all tested data sets, the resulting model can even be competitive with BERT. A further advantage is that the inference time doesn't depend on the amount of training data examples, as is the case with KNN-based approaches.
 \end{enumerate}

 Both findings apply to `low-resource' few-shot setups as well as to scenarios with many training data.

\bibliographystyle{plain}
\bibliography{references}


\begin{appendices}
\section{Expected accuracy}
\label{app:accuracy}
Let us first view how we can calculate expected accuracy of KNN in the general case with any $K$. Here, we understand as the expected accuracy the accuracy that we achieve on average if we use a random guesser to break ties. 

Consider $c^t$ the number of candidates in a tie $t$. So, on a data set with $N$ untied examples and $T$ tied examples, we have
\begin{equation}
E[Accuracy]= \frac{\sum_{i=1}^N I[pred(i) = label(i)] + \sum_{t=1}^T 1/c^t}{N+T},
\end{equation}
where $I[x]$ returns 1 if $x$ is true and and 0 else. When $K=2$, we have $c^t=\frac{1}{2} ~\forall t$, i.e.,

\begin{align}
    & \frac{\sum_{i=1}^N I[pred(i) = label(i)] + T/2}{N+T} \\
    =~& \frac{1}{2}\bigg(\frac{\sum_{i=1}^N I[pred(i) = label(i)] + T}{N+T} + \frac{\sum_{i=1}^N I[pred(i) = label(i)]}{N+T}\bigg) \\
    =~& \frac{1}{2}\bigg(O + P\bigg), 
\end{align}
\end{appendices}

with $O$ the optimistic accuracy and $P$ the pessimistic accuracy.

\section{Hyper-parameters of \simple}

See Table \ref{tab:filter_ablation}. All presented KNN numbers are expected (realistic) accuracy scores of using  K=2 and random tie-breaking.
\label{app:hyper-param}
\begin{table}
    \centering
    \scalebox{0.82}{\begin{tabular}{lrrrrr}
    \toprule
    variant & 20News &  KinyarwaNews & KirundiNews	& (DengueFilipino) & SwahiliNews \\ 
    \midrule
       \simple (n=3) & 0.578 & 0.825 & 0.840 & \textbf{0.885} & 0.834 \\   
       \simple  n=0 & 0.574 &\textbf{ 0.829} & 0.838 & 0.865 & 0.830 \\
       \simple  n=1 & 0.571 & 0.827 & \textbf{0.856} & 0.865 & 0.832 \\
       \simple  n=2 & \textbf{0.579} & 0.826 & 0.841 & 0.876 & 0.834 \\
       \simple  n=4 & 0.571 & 0.826 & 0.851 & 0.879 & 0.832 \\
        \simple n=10 & 0.411 & 0.734 & 0.770 & 0.418 & 0.565 \\ 
        \simple keep only $n\leq 3$ & 0.360 & 0.661 & 0.839 & 0.834 & 0.545 \\
       \simple  only split, n=3 & 0.550 &  0.826 & 0.851 & 0.871 & 0.833 \\
       \simple  only split, n=0 & 0.538 & 0.829 & 0.838 & 0.842 & 0.835 \\
        \midrule
        \simpletr (tfidf)& \textbf{0.853} & \textbf{0.889} &  \textbf{0.899} & 0.991& \textbf{0.899} \\
        \simpletr inary &  0.799 & 0.874 & 0.891 & \textbf{0.997} & 0.875 \\
        \simpletr count & 0.786 & 0.864 & 0.879 & \textbf{0.997}& 0.879 \\
        \bottomrule
    \end{tabular}}
    \caption{\simple variants, removing tokens with length $\leq$ n, and abstaining from any string normalization (only split, i.e., no removal of dots and commas, no lower-casing), with and without removing tokens that do not exceed a certain length. Full training data.}
    \label{tab:filter_ablation}
\end{table}

\end{document}